\documentclass[10pt, conference, compsocconf]{IEEEtran}
\ifCLASSINFOpdf
\else
\fi
\usepackage[cmex10]{amsmath}
\usepackage{bbm}
\usepackage{graphicx}
\usepackage{latexsym}
\usepackage{comment}
\usepackage{bbm}
\usepackage{ mathrsfs }
\usepackage{ amssymb }

\usepackage{epstopdf}
\usepackage{url}

\begin{document}
%
\title{Optimizing Waiting Thresholds Within A State Machine}


\author{\IEEEauthorblockN{Rohit Pandey}
\IEEEauthorblockA{Azure Compute Insights\\
Microsoft\\
Redmond, WA\\
rohitpandey576@gmail.com}
\and
\IEEEauthorblockN{Yifan Chang}
\IEEEauthorblockA{Azure Compute Insights\\
Microsoft\\
Redmond, WA\\
yifan.chang@microsoft.com}
\and
\IEEEauthorblockN{Cameron White}
\IEEEauthorblockA{Azure Fabric Controller\\
Microsoft\\
Redmond, WA\\
jamw@microsoft.com}
\and
\IEEEauthorblockN{Gaurav Jagtiani}
\IEEEauthorblockA{Azure Fabric Controller\\
Microsoft\\
Redmond, WA\\
gajagt@microsoft.com}
\and
\IEEEauthorblockN{Aerin Young Kim}
\IEEEauthorblockA{Cloud AI\\
Microsoft\\
Redmond, WA\\
aerinykim@gmail.com}
\and
\IEEEauthorblockN{Gil Lapid Shafriri}
\IEEEauthorblockA{Azure Compute Insights\\
Microsoft\\
Redmond, WA\\
gilsh@microsoft.com}
\and
\IEEEauthorblockN{Sathya Singh}
\IEEEauthorblockA{Azure Fabric Controller\\
Microsoft\\
Redmond, WA\\
sathyasi@microsoft.com}
}


%


\maketitle

\begin{abstract}
Azure (the cloud service provided by Microsoft) is composed of physical computing units which are called nodes. These nodes are controlled by a software component called Fabric Controller (FC), which can consider the nodes to be in one of many different states such as Ready, Unhealthy, Booting, etc. Some of these states correspond to a node being unresponsive to FCs requests. When a node goes unresponsive for more than a set threshold, FC intervenes and reboots the node. We minimized the downtime caused by the intervention threshold when a node switches to Unhealthy state by fitting various heavy-tail probability distributions. 

We consider using features of the node to customize the organic recovery model to the individual nodes that go unhealthy. This regression approach allows us to use information about the node like hardware, software versions, historical performance indicators, etc. to inform the organic recovery model and hence the optimal threshold. In another direction, we consider generalizing this to an arbitrary number of thresholds within the node state machine (or Markov chain). When the states become intertwined in ways that different thresholds start affecting each other, we can’t simply optimize each of them in isolation. For best results, we must consider this as an optimization problem in many variables (the number of thresholds). We no longer have a nice closed form solution for this more complex problem like we did with one threshold, but we can still use numerical techniques (gradient descent) to solve it.

\end{abstract}


%
\IEEEpeerreviewmaketitle

\section{Introduction}
Section 2 will briefly go over the architecture of Azure and the data we use for modeling. Sections 3 and 4 will formulate the problem mathematically. In section 5, we will discuss extending the model to multiple thresholds and in section 6, we will explore ways to use regression, so we can leverage features to customize our recovery models.

Azure consists of many nodes, which are the smallest computational hardware units that hosts several virtual machines. To be able to deliver promised uptime service level agreements (SLAs) to customers, Azure must keep the nodes operational. Many different factors can cause downtime, such as planned maintenance, system failures, communication failures, etc. The component that causes downtime also determines what the effect of downtime is and how much it costs. For example, in the case of a virtual machine (VM) failure, only one VM is affected, however, in the case of a node failure all the VMs that are hosted on that node are affected. To meet promised SLAs to customers, Azure should keep downtime caused by higher level components to a minimum. One of the most important components that is prone to failure is the node.

In the production setting, Azure uses a software component called Fabric Controller (FC) to monitor and control the states of nodes. Nodes can be in one of the many different states; and they spend most of their time in Ready state, where the FC can communicate with them. At times, a node can switch to the Unhealthy state, which means that the node has not responded to FCs requests for a set amount of time. This situation can happen under different circumstances, such as heavy computational load, software failure, hardware failure, etc. In these cases, to reduce downtime that is caused by an unhealthy node, FC can intervene and decide to take different actions or can decide to wait for node to recover. Depending on the history of node, FC can mark it for human investigation, decommission, or power cycle. However, all of these actions cost Azure downtime. Here, we analyze the cost of waiting for spontaneous recovery and the cost of intervening to find an optimum threshold for intervention. To do so, we develop an analytical model describing the spontaneous recovery and calculate how much downtime Azure incurs with intervening. We describe in detail the analytical model and a robust implementation of the fitting of the model to the data (including regressing with features of the node to customize the organic recovery distribution per node). Finally, we show that moving the current intervention threshold from 10 minutes to its optimum value can help us reduce the total downtime by 10\%. 

For regressing the survival data on features, we found that while there is substantial literature on regressing survival data (see [1] and [2]) such as the Cox model, the emphasis tends to be on interpretability. So, the methods discussed generally end up regressing just one of the parameters of the distribution used to model the survival (like scale parameter) instead of simply regressing all of them. We found that for our purpose, the latter worked better since performance (in terms of downtime savings) and not interpretability was the major concern for us. Further, we didn’t find any prior work that combines survival analysis with the Markov theory for state transitions to reach optimal decisions.

\begin{figure}
\includegraphics[width=\linewidth]{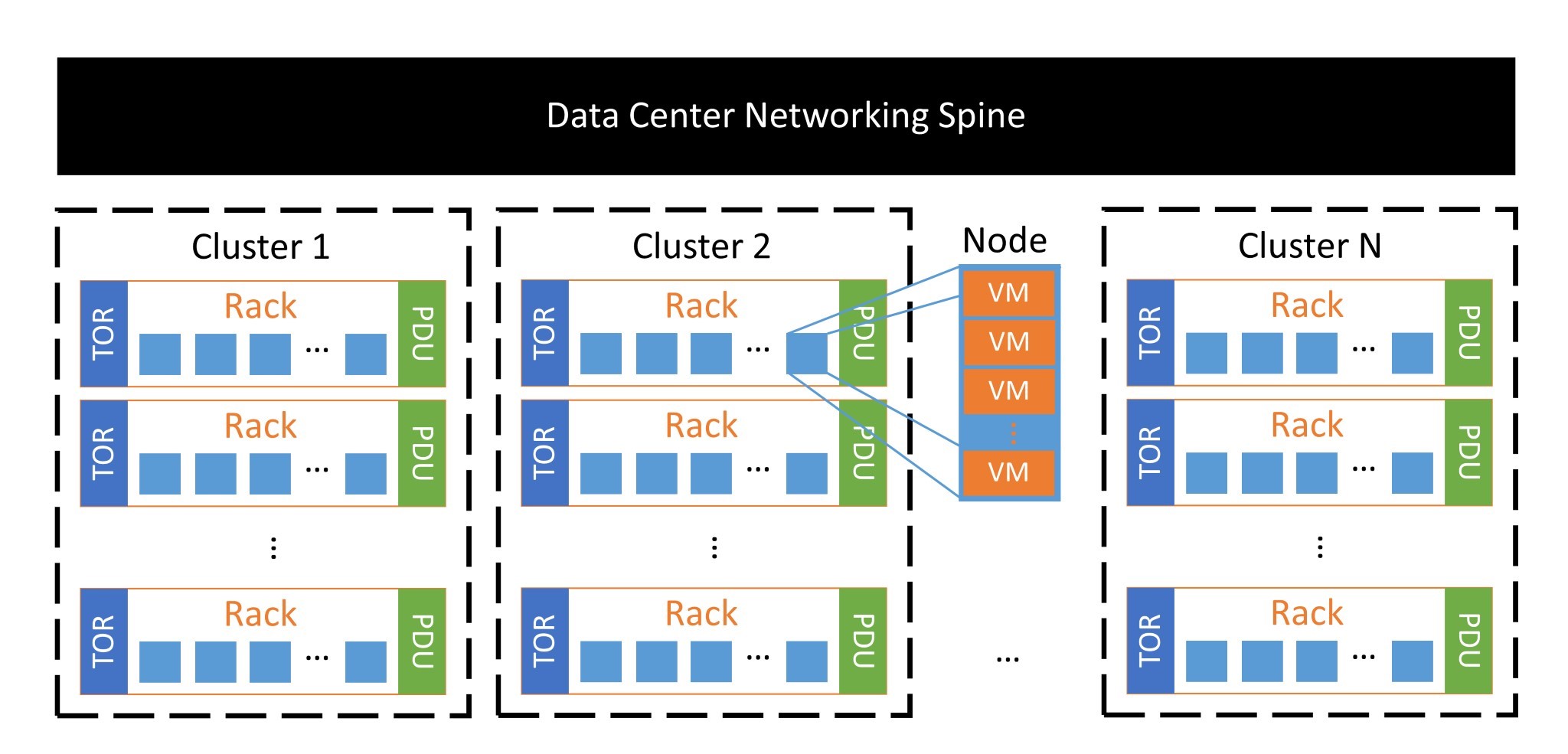}
\caption{A brief illustration of Azure architecture.}
  \label{fig:fig1}
\end{figure}

\section{Brief overview of Azure and dataset }
Azure has several hardware and software layers that allows management of resources to be easier. In this section, we explain the layers of Azure architecture related to the problem in hand briefly. The basic building blocks of Azure are virtual machines (VMs). Nodes, also called blades, host VMs that are the smallest constituent unit of the hardware stack. A number of nodes together sit in a rack, which also contains a power distribution unit (PDU) and a top-of-rack switch (ToR). A collection of racks make a cluster and a group of clusters residing in the same building are called a Datacenter. We show an illustration of this in Figure 1. Every cluster has its own FC, which controls the nodes in that cluster through direct connection and PDUs in the case that a direct connection is not possible. FCs make decisions on what to do with nodes at their disposal, i.e. provisioning new VMs, updating node’s software, etc. To be able to manage nodes, FC keeps track of all node states. From FCs viewpoint, there are twenty-two different states that a node can be in. Only seven of these states are relevant to the intervention of an unhealthy node by FC. Here we give a brief overview of these states:

\begin{itemize}
\item{Raw: Default state. A node is marked Raw when FC does not know about its state. This state is reached after FC fail-over, a new node turning on for the first time, etc.}

\item{Ready: A proxy for an operational state. A node is marked Ready, when FC can communicate with it.}

\item{Unhealthy: A node is marked Unhealthy, if FC cannot communicate with it for more than a minute. }

\item{Booting: Unoperational state. A node is marked Booting, when intervention threshold is reached without communication from the node and there is an entry in PXE (pre-execution booting environment) request list coming from the node, indicating that the node is in the process of rebooting (soft reboot). }

\item{PoweringOn: Unoperational state. A node is marked PoweringOn, when intervention threshold is reached, and the node is power cycled by FC through the PDU. This process is equivalent to pulling the plug on the machine and then turning it on again (hard reboot). }

\item{HumanInvestigate: Unoperational state. A node is marked HumanInvestigate, when FC could not make the node operational again. In this case, node is handed off to Datacenter Manager (DCM) for further investigation. VMs on a node marked HumanInvestigate are moved to another node. }

\end{itemize}

Every time a node switches from one state to another, FC logs an entry detailing the duration spent in the former state. From these event logs, we can create a distribution of durations for each state transition. As mentioned earlier in Section 1, FC intervenes if a node spends more than a set threshold in Unhealthy state, which is currently set to 10 minutes uniformly across all Azure due to historical reasons. If a node switches to Ready state from Unhealthy without intervention, we assume that it spontaneously recovered. Therefore, we are mostly interested in the transitions from Unhealthy to Ready, and Unhealthy to Booting or PoweringOn. In Figure 2, we provide a graph of relevant states and the transitions among these states. These transitions can be summarized in terms of matrices of transition probabilities and transition times. These matrices will be described in detail in the subsequent sections. 

\begin{figure}
\includegraphics[width=\linewidth]{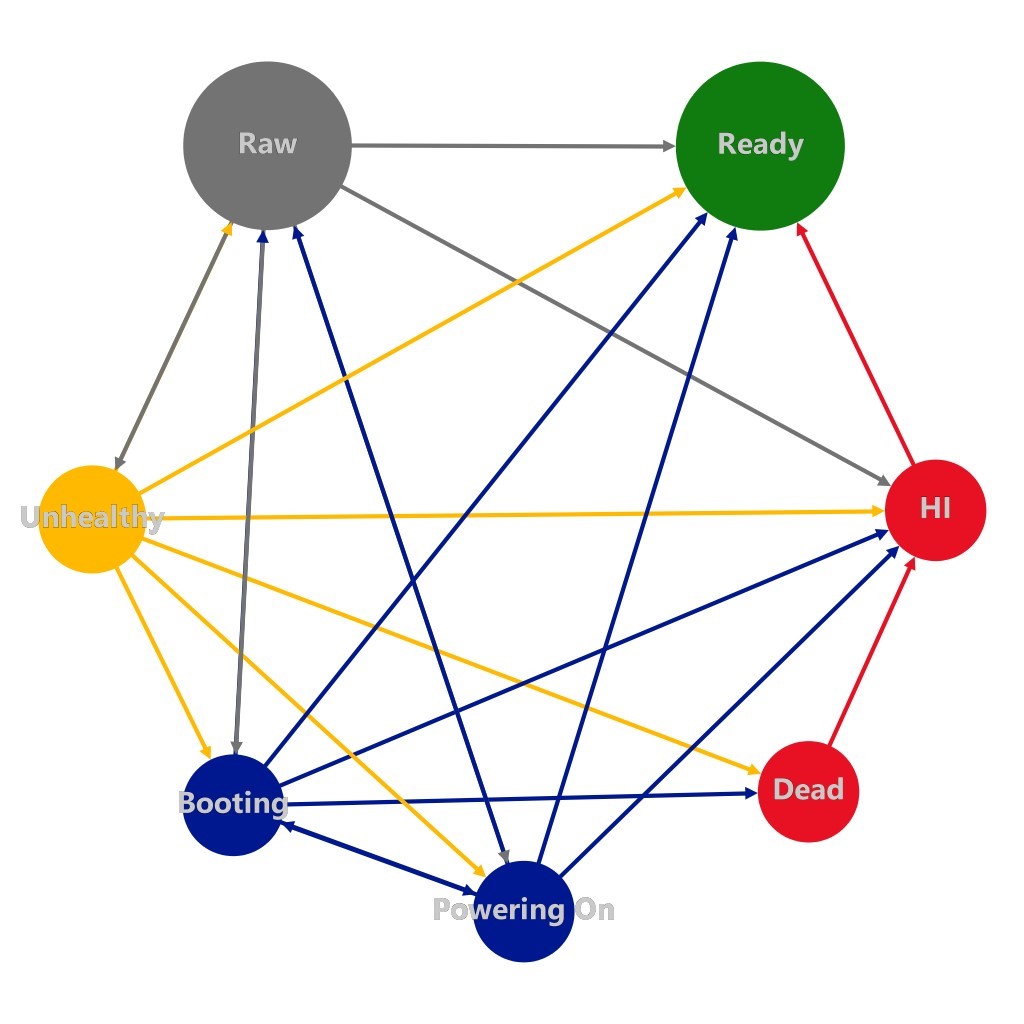}
\caption{State transition graph of relevant states.}
  \label{fig:fig2}
\end{figure}

\section{Analytical formulation of the problem}
In this section, we formulate the problem of determining the optimum threshold analytically by building our objective function. Let’s consider the downtime a node will suffer (which we will represent by DT). DT will depend on the time it takes for the node to organically recover (i.e. go from Unhealthy to Ready, $T$). If the recovery takes less than the threshold that the FC waits ($\tau$), no reboot action will be taken and DT will just be the organic recovery time $T$.  Otherwise, we would have already waited the threshold amount of time ($\tau$) and will need to wait a further time required to reboot the node, represented by $C_{int}$.

\begin{equation}
    DT=
    \begin{cases}
      T, & \text{if}\ T < \tau \\
      (\tau + C_{int}), & \text{otherwise}
    \end{cases}
 \end{equation}

As mentioned in Sec. 1, we choose the expected downtime per node as our metric, which we can now define as (weighted sum of the expected values from the two scenarios with the probabilities of the two scenarios):

\begin{equation}
 \mathbbm{E}[DT] =  P(T \leq \tau)\mathbbm{E}[T | T < \tau] +   P(T > \tau)(\tau + C_{int}), \label{eq:expectedDT}
\end{equation}

To summarize: 

\begin{itemize}
\item{$T$ describes the organic distribution of the node recovering, which can be approximated by various distributions for a node going from Unhealthy to Ready.}
\item{$\tau$ is the intervention threshold that FC waits on an Unhealthy node before taking action.}
\item{$C_{int}$ is the time it will take for a node to get to Ready state after intervention (although this is random, the variance is generally not high hence it is safe to assume that it does not depend on τ).}
\item{$DT$ is the downtime a node will incur under a given intervention threshold ($\tau$), it is a random variable that depends on $T$ and $\tau$.}
\end{itemize}

Now,
 \begin{equation}
\mathbbm{E}[T | T < \tau] = \frac{\int_0^\tau t f_T(t)dt}{P(T < \tau)}, \label{eq:expectation}
\end{equation}

where $f_T(t)$ is the probability density function of variable $T$. By substituting equation 3 in equation 2, we get: 

\begin{equation}
\mathbbm{E}[DT] = \int_0^\tau t f_T(t) dt + [1 - P(T \leq \tau)] \times [\tau + C_{int}]. \label{eq:subDT}
\end{equation}

To find the optimum τ that minimizes $E[DT]$, we take the derivative of equation 4 and set it to zero:

\begin{subequations}
\begin{align}
0 &= \frac{dE[DT]}{d\tau}, \\
0 &= \tau f_T(\tau) - f_T(\tau) \times [\tau + C_{int}] + [1 - P(T \leq \tau)], \\
1 - P(T \leq \tau) &= f_T(\tau)  C_{int}, \\
\frac{f_T(\tau)}{P(T \geq \tau)} &=  \frac{1}{C_{int}}\label{eq:hazard}.
\end{align}
\end{subequations}

Eq. 5d has a very intuitive meaning. The left-hand side is called the \textit{Hazard Rate} of distributions describing the arrival times of certain events. It is the probability we will witness an event in the next unit of time conditional on not witnessing it until now.

 The events being modeled are generally negative in nature hence the \textit{hazard} in the name. Nevertheless, in our case, the event we anticipate is positive, namely node going back to \textit{Ready} state. Here, the rate is described as the inverse of the average time until the next event arrives as seen from the current state, instantaneously. Note that for most distributions, this rate itself will change with time so the average time until the next event won’t actually be the inverse of the instantaneous rate, i.e. the instantaneous velocity of an accelerating object at a certain time cannot on its own predict the time the object will reach a certain point. The right-hand side is the inverse of the (deterministic) time it takes for a node to go back to \textit{Ready} after intervention.  We can see that this is a kind of rate as well since we get exactly one recovery in $C_{int}$ amount of time, so the recoveries per unit time is $\frac{1}{C_{int}}$. Hence, the optimum $\tau$ is achieved when the rates corresponding to the two competing processes are equal. Now, all that is left is to pick a good distribution for $T$ and solve equation (5d). We will shop for such candidates in the next section.

Let’s consider how the expected downtime described in equation (4) varies with our unhealthy threshold $\tau$.

When $\tau=0$, we will always immediately reboot the machine. The downtime incurred will always be $C_{int}$. 

When we will never reboot any machine. We will just wait indefinitely. The expected downtime in this scenario will just be the expected value of the distribution describing the organic recovery ($E(T)$). Note that for some distributions (like Lomax when its shape parameter is more than one), the expected value blows up (tends to infinity). In this case,  as  $E[DT] \to \infty $ as $\tau \to \infty$. 

We showed in equation (5d) that there will be certain values of $\tau$ where the hazard rates of the organic and inorganic recoveries align where $E[DT]$ will have optima. For distributions with monotonically decreasing hazard rates (like the Lomax and Weibull – with shape parameter less than one), this will be exactly one minima. 

Using these three conditions, we end up with the plot showing the relationship between $E[DT]$ and $\tau$ depicted in Figure 3 (it mentions an estimate for the downtime savings which we will discuss further in the results section). 

\begin{figure}
  \includegraphics[width=\linewidth]{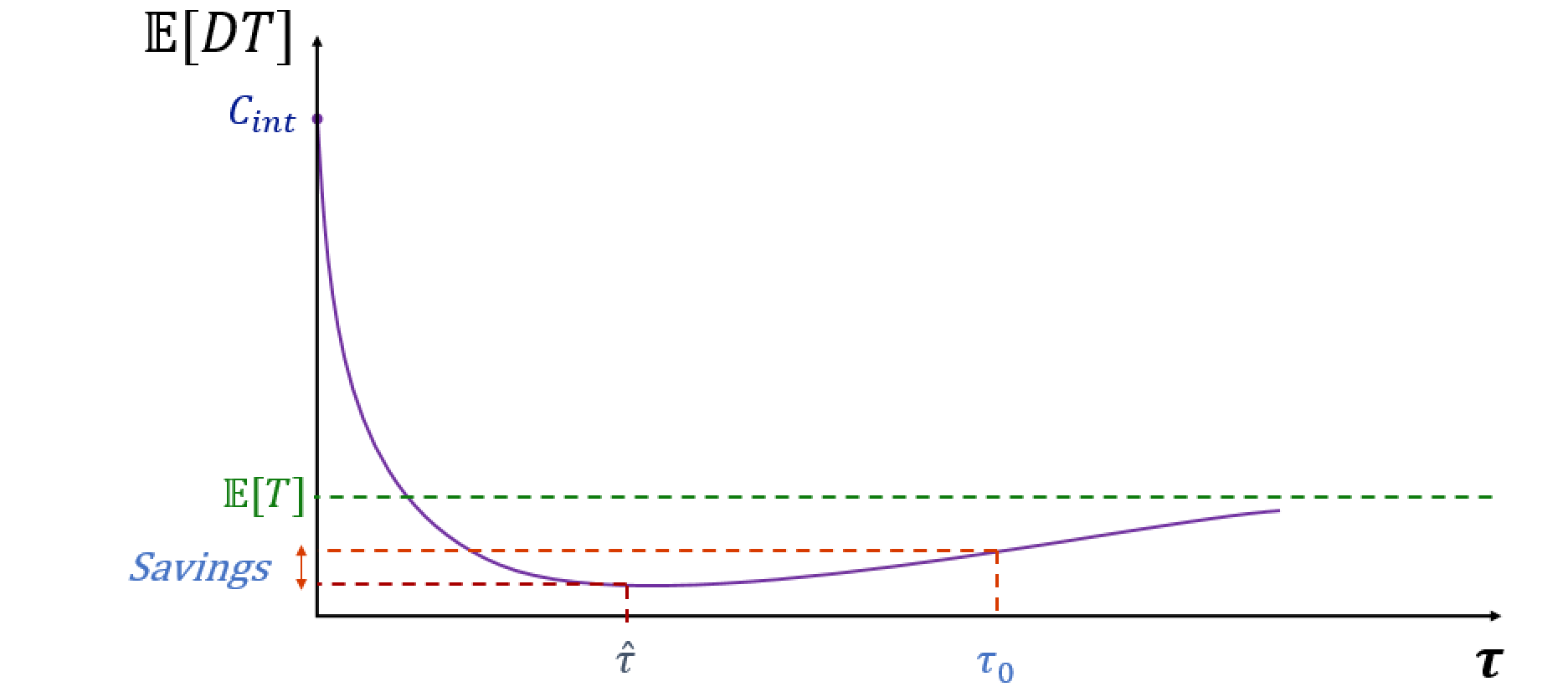}
  \caption{Optimum intervention threshold and relative savings. The x-axis is the waiting threshold (amount of time to wait for an Unhealthy node to recover) and the y-axis is the expected downtime the node will incur. The green horizontal line is the expected value of the organic recovery distribution, the red line is the minimal downtime we can possibly incur and the orange line is the amount of downtime incurred at the current threshold.}  
\end{figure}

Note that the flat profile for $\mathbbm{E}[T]$ around the optimal threshold suggests to us that there might be scope for adding a term to the objective function that penalizes any single node staying down for too long. That way, we will be able to fine-tune our value of the optimal $\tau$ from a neighborhood where the expected downtime is practically the same. The authors of \cite{apaar_paper} explore such objective functions. In this paper however, we will stick to minimizing the expected downtime.

\subsection{Criterion for choosing the distribution to model the organic recovery, T}
Having defined the problem at a broad level, the next step is to get into the details and pick a good distributional assumption for the organic time to go from the Unhelathy to the Ready state. We will enumerate some properties we would want such a distribution to possess. From the discussion in the last section, we know that the Hazard rate of the candidate distributions is a critical property. Also, we know from the data on the organic transitions between Unhealthy to Ready provided by the Fabric Controller team that the distribution is increadibly heavy tailed. 

We got an intuitive sense from the data and our discussions with the Fabric team that as time passes, the probability that a given node will come up on its own any time soon decreases drastically. We can interpret this as the Hazard rate of the distribution decreasing with time (new events of the node going back to Ready become increasingly sparse). However, as we slice and dice the data, we end up with many possible profiles for the hazard rate (monotonically decreasing, increasing to a mode before decreasing, etc.). 

Here are some properties we would want our distribution to possess in decreasing order of importance:

\begin{enumerate}
\item{The support of the distribution should over $[0,\infty)$.}
\item{This distribution should be flexible enough to model all kinds of hazard rate profiles. In particular, decreasing and increasing to a mode before decreasing.}
\item{The Hazard rate should tend to zero as the time waited tends to infinity. This ensures that the expected downtime will have a minima (as we expect) as long as the Hazard rate begins above $\frac{1}{C_{int}}$.}
\end{enumerate}

We now evaluate some distributions we considered and see how they stack up. All distributions considered here satisfy property (1). 

\begin{enumerate}
\item{Exponential: This does not satisfy property (2) since the Hazard rate is constant with time. In essence, this means the probability of the node going to ready in the next 10 minutes is the same regardless of how much time we have already waited. This is undesirable behavior as we know that greater time spent in the Unhealthy state should decrease the probability of coming back up. Note that this distribution will have us wait an infinite time if its hazard rate is higher than $\frac{1}{C_{int}}$ and wait no time otherwise before rebooting since it's hazard rate is constant.}
\item{Weibull distribution: This distribution can only model monotonically increasing or monotonically decreasing hazard rates. It is also not very good at modelling heavy tails.}
\item{Lomax distribution: This distribution can model only monotonically decreasing hazard rates. The decrease in the Hazard rate with time ($t$) is of the order $O(t^{-1})$.}
\item{LogLogistic distribution: This distribution is very flexible and can model all kinds of hazard rate profiles. It is also good at modeling heavy tails.}
\end{enumerate}

In the next section, we take the Lomax distribution as a template (it makes for a good representative since it is analytically tractable) and evaluate some important properties of this distribution including the expected downtime as a function of the threshold $\tau$ we defined in equation (2). The process for using other distributions will be very similar apart from the fact that some of the equations might need numerical methods for solving.

It should be noted however, that the Lomax distribution can model only monotonically decreasing profiles for the hazard rate. This happens to be desirable on the global level but proves too restrictive when we try start slicing the data by various features (properties of the node like hardware model, software version, etc.). 

We might find for example, that in most instances, nodes with a certain hardware take between five and ten minutes to organically recover as opposed to a monotonic distribution. In other words, the mode might be at a positive value instead of at zero. The Lomax distribution can only model the latter profile. In such cases, the LogLogistic distribution proves more versatile since it is able to model both monotonically decreasing and non-zero mode profiles and satisfies other requirements. Figure 4 below shows the various possible profiles the organic recovery might follow and the corresponding hazard rates. The third column of the figure shows the corresponding expected downtime to waiting threshold profile (with the very first one being the same as figure 4). 

\begin{figure}
  \includegraphics[width=\linewidth]{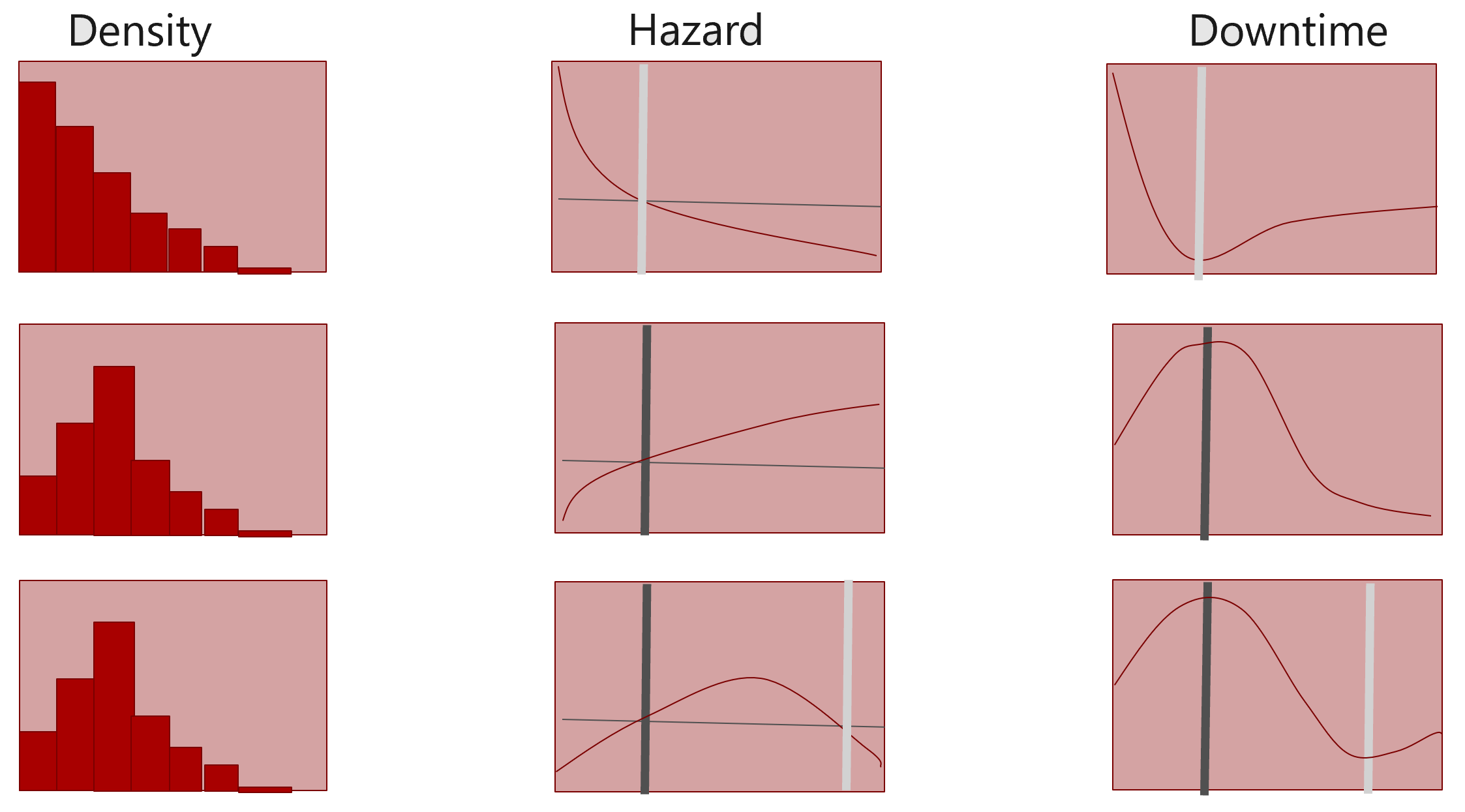}
  \caption{Various possible profiles the organic recovery can follow (first column); the corresponding hazard rates (second column) and behavior of expected downtime with threshold (third column). This light grey lines represent minima where we would want to set our optimum threshold while the dark grey lines are the maxima.}  
\end{figure}

\subsection{Application of the Lomax distribution to model organic recovery}
The Lomax distribution is a shifted version of the more famous Pareto distribution (famed for having a heavy tail which decays polynomially) so that the support is over $[0,\infty)$ as we desire. It has two parameters, the shape parameter $\kappa$ and the scale parameter $\lambda$. We will now list some of its properties as we will use them frequently in the forthcoming sections.

The probability density function is given by - 

\begin{equation}f_X(x) = \frac{\lambda.\kappa}{(1+\lambda.x)^{\kappa+1}}\end{equation}

The cumulative density function (CDF - probability that a sample is less than a certain value, $x$) is given by - 
\begin{equation} F_X(x) = \int_0^x f_X(t) dt = 1- (1+\lambda.x)^{-\kappa}\end{equation}

The 1 - the CDF is the survival function which represents the probability that a sample will be greater than a given value $x$ - 
\begin{equation}\mathcal{S}_X(x) = (1+\lambda.x)^{-\kappa} \end{equation}

Using the density and survival functions from (7) and (8), we can calculate the all important Hazard function defined in equation (2) - 
\begin{equation}\mathcal{H}_X(x) = \frac{\lambda.\kappa}{1+\lambda.x} \end{equation}

We can now use the Hazard function in equation (9) to find the optimal value of the parameter $\tau$ - 

\begin{equation}\hat{\tau} = \frac{C_{int}}{\kappa} - \frac{1}{\lambda}\end{equation}

Another important property somewhat related to the Hazard rate is the expected value of the Lomax random variable given that it is already greater than a certain value, $x$. Since the rate of arrival of events (hazard rate) decreases as the inverse of a linear function, we should expect the average time to the next arrival to increase linearly, which it indeed does. You can find a derivation of this quantity in Appendix A.
\begin{equation} \mathbbm{E}[X|X>x] = \int_x^\infty t f_X(t) dt = \frac{1}{\lambda.(\kappa - 1)} + \frac{\kappa}{\kappa-1} .x\end{equation}

Substituting $x=0$, we get the expected value - 
\begin{equation}\mathbbm{E}[X] = \frac{1}{\lambda (\kappa - 1)} \end{equation}

Using the properties above, we can evaluate the expected downtime $T$ given a threshold $\tau$ as defined in equation (1). A derivation of this expression is given in Appendix B. The value of $\tau$ that minimizes this function is already given in equation (7).
$$ \mathbbm{E}(T) = \left( \frac{1}{\lambda.(\kappa - 1)} - \left[ \frac{1}{\lambda.(\kappa-1)} + \tau.\frac{\kappa}{\kappa-1}\right].(1+\lambda.\tau)^{-\kappa}   \right) $$
\begin{equation} + \left((\tau+C_{int}).(1+\lambda.\tau)^{-k}\right) \end{equation}

\subsection{Estimating the parameters of the Lomax distribution}
The usual formulation of the Maximum Likelihood problem for a certain distribution with probability density function (pdf) $f_X(x)$ given some samples $x_i$ where $i \in [1,2, \dots N]$ involves a maximization of the Likelihood function (the likelihood of observing the given samples assuming certain parameters $\Theta$ for the distribution) - 

$$ \mathscr{L}(\Theta; x_1, x_2, \dots x_n) = f_X(x_1, x_2, \dots x_n | \Theta) = \prod_{i=1}^n f_X(x_i | \Theta)$$

This works well for the samples in our data for the transition from Unhealthy to Ready. However, this is not the only information we have. We also know how many times nodes were taken from the Unhealthy to Booting or PoweringOn states. And in these instances, we know they must not have come up for at least $\tau_0$ seconds (the current threshold that the data we get from the Fabric Controller team is based on). Lets say there were $m$ such instances in the data. For these, we don't know the exact times they would have taken to go to Ready since Fabric intervened before they could. So, their times to go to ready must have been greater than $\tau_0$. The probability of a sample of a distribution being above a certain value is described by the Survival function $\mathcal{S}(x)$ (the expression for which we provided for the Lomax distribution in the previous section). With this in mind, we modify our likelihood function to account for these $m$ samples by further multiplying by the survival function at the $\tau_0$.

$$ \mathscr{L}(\Theta; x_1, x_2, \dots x_n, m) =  \left(\prod_{i=1}^n f_X(x_i | \Theta)\right). \left(\prod_{j=1}^m \mathcal{S}(\tau_0|\Theta)\right)  $$
$$ =  \left(\prod_{i=1}^n f_X(x_i | \Theta)\right).  \mathcal{S}(\tau_0|\Theta)^m $$

Taking the logarithm on both sides, we get the more tractable and numerically stable log-likelihood function - 
\begin{multline} \ell(\Theta;x_1,x_2,\dots, x_n,m) = \left(\sum_{i=1}^{n}\log\left(f_X(x_i|\Theta)\right) \right) \\+ m.\log(\mathcal{S}(\tau_0|\Theta)) \end{multline}

Substituting the values of the pdf and survival functions from the previous section (equations (3) and (5)) into the log-likelihood function above we get - 

\begin{multline} \ell(\kappa,\lambda) = n.\log(\kappa.\lambda) \\- (\kappa+1) \sum_{i=1}^{n} \log(1+\lambda.x_i) - m.k.\log(1+\lambda.\tau_0)\end{multline}

We can now differentiate this function with respect to $\lambda$ and $\kappa$ and this leads to the following conditions - 

\begin{equation} \hat{\kappa} = \frac{n}{\sum_{i=1}^n \log(1+\hat{\lambda}.x_i) + m.\log(1+\hat{\lambda}.\tau_0)}\end{equation}

\begin{equation} \frac{n - m \hat{\kappa} \frac{\tau_0} {1+\hat{\lambda}\tau_0}}{\hat{\lambda} \left(\sum_{i=1}^n \frac{x_i}{1+\hat{\lambda}.x_i} \right)} -1 = \hat{\kappa}  \end{equation}

Equation (15), once its R.H.S is substituted with (14) has just one variable, $\hat{\lambda}$. Hence, it can be solved with a simple method such as bisection. And then, it can be plugged into equation (14) to find $\hat{\kappa}$. More details are in [3].

\section{Estimating the cost of intervention with Markov Chains}
Another crucial aspect of the problem is the cost of intervening a node ($C_{int}$) as stated in Eq. 5d. In many cases where the threshold is crossed, nodes go to PoweringOn state from Unhealthy. In the simplest case, they will simply reboot and go to Ready state. Nonetheless, there are many other paths node can take before reaching Ready state. For example, a node can go from PoweringOn to HumanInvestigate in the case of software and/or hardware problems and then the customer VMs are taken off it before it finally gets back to Ready. How can one account for all these different paths to come up with a single expected value for $C_{int}$? 

We can model this process of transitioning between states as a Markov Chain. Markov chains are used to model probabilistic processes, where the next state of the system being described depends only on the current state and not the history of previous states visited. We denote the probabilities of going from a given state i to another state j in a single transition (i.e., PoweringOn to HumanInvestigate) by $P_{i,j}$. We can estimate this by dividing the number of transitions from state i to j dividing by the total transitions from state i. Hence, we get $P_{n,m}$ transition matrix describing the probabilities of moving between states. Note that the rows of $P$ sum to 1. At the same time, there is an average downtime cost associated with transitioning from state i to state j and we denote this by $T_{i,j}$. Thus, we get another matrix, $T_{n,n}$. We periodically calculate these matrices (once a day) based on a rolling window of the past 30 days. This way, we can log their values in a data stream and keep an eye on how stable their values are. As one might expect due to the large period of aggregation, we found them to be quite stable, moving not more than 1\% between most days. Now, let’s look at how we can use these matrices to estimate the average time a node needs to go from Unhealthy to Ready (or indeed, any two states). 

We define the state Ready as the “absorbing state”, such that $P_{Ready,i}=0$ and $T_{Ready,i}=0$. The other states are called “transient” states. We also define vector $t$, where $t_i$ gives the average duration to arrive to Ready state from state i. We can now describe a system of equations for finding $t_j$. 
When in state i, let’s say we first move to state j. This process takes time $T_{i,j}$ And once in state j, we still need to get to the absorbing state (Ready). And by definition of the vector t, the time taken to go to Ready from state j will be $t_j$. So, if the node goes first to 𝑗 from 𝑖, the total time to get to Ready turns out to be $T_{i,j}+t_j$. Note that if j happens to be Ready, then $t_j$ is zero. 

But, the probability of going to state j is $P_{i,j}$. So, for the expected value, we need weigh the time spent in this path by it. And similarly, we will need to do this across all possible values of j. Doing this, we get the following system of equations.

\begin{subequations}
\begin{align}
t_i &= \sum_{j=1}^{n} P_{i,j}(\mathcal{T}_{i,j} + t_j),\\
t_i - \sum_{j=1}^n P_{i,j} t_j &= \sum_{j=1}^n P_{i,j} \mathcal{T}_{i,j}.
\end{align}
\end{subequations}

This is a system of ($n - 1$) linear equations since $t_{Ready}=0$. It can be stated succinctly in matrix form as:

\begin{equation}
(\mathbbm{I} - \mathbf{Q})\mathbf{t} = (\mathbf{P} \odot \mathbf{\mathcal{T}}) \cdot \mathbf{1}, \label{eq:cost}
\end{equation}

where $\mathbbm{I}_{n-1 \times n-1}$ is the identity matrix, $Q_{n-1 \times n-1}$ is the transition matrix of transient states where the rows sum to less than 1 since Ready column is removed from $P$, $\odot$ is the Hadamard product of matrices where two same size matrices are multiplied element-by-element, $\mathbf{1}_{n-1}$ is a vector of ones. 

After solving Eq. (18), we use $t$ to find the average time it takes a node from intervening states to Ready. This will give an estimate for $C_{int}$ , which we then use in Equation (5d). Notice that we implicitly assumed here that $C_{int}$ will not depend on $\tau$ itself, which might not always be a good assumption (though it is for our data). We generalize this framework to account for this possible dependency in section 8.

\section{Modelling organic recovery}

From the previous sections, we know that apart from describing the motion of the nodes between different states of the transition matrices, the other important aspect that affects the model is the distribution of the time it takes for organic recovery. This organic recovery describes the time taken for the node going from certain states to Ready. In particular, Unhealthy to Ready, Booting to Ready and PoweringOn to Ready. When the node is Unhealthy, there is the time $T_1$ that it will take to spontaneously go to Ready. When it is PoweringOn, there is another time $T_2$ it will take to go to Ready and when the node is in Booting, there is a similar time, $T_3$ for it to go from Booting directly to Ready. So far in this paper, we have focused on the Unhealthy->Ready scenario. 

The important question is, how do we estimate the distributions of these $T_i$’s. These distributions play crucial roles in the transition matrices and by extension, the optimal thresholds, so it is of utmost importance that we model them as closely as possible to reality. On the face of it, it should be best to use non-parametric models (ones that assume no functional form but simply follow the raw data) so we’re not in any danger of working with the wrong distributional assumption. On closer reflection however, we find that there are certain advantages to working with parametric models – 

\begin{itemize}
\item{	Our data for the recovery profiles of the $T_i$s has been censored at some level since the inception of Azure. This means we never get to observe the uninhibited samples from these distributions. At some point, the samples are just cut off at a certain threshold and all we can say is that the recovery took longer than the said threshold for those samples. As the thresholds are lowered, this censoring problem gets worse. Non-parametric models rely on the raw data. So, they can’t deal with this censoring. Parametric models however, can extrapolate their distributional assumptions to regions where there have been no samples. }
\item{With parametric models, we can use an arbitrary number of features associated with the node going into a particular state to customize the recovery distribution. Since non-parametric models use raw data, the only way to customize them to a feature set is by splitting the data across possible values of the feature set and training the models separately. This becomes problematic when the number of features under consideration is too large since the data gets spread too thin the various combinations. With parametric models however, we can simply express the parameters as a function of the features and use regression techniques. We will discuss how we might go about doing this in the current section. }
\end{itemize}


\subsection{Using features to customize the recovery profile for each data point}
Before we dive into how we would create a regression model for modelling organic recovery, let’s start with a simple organic recovery model without any features. The traditional likelihood function for a $T$ which follows the PDF - $f_T (t)$ and parameter vector $\theta$ (for example, for a Lomax distribution, $\theta$ would consist of the shape ($\kappa$) and scale ($\lambda$) parameters from equation (6)) is given by –

\begin{equation}
L(\theta; t_1, t_2, \dots t_n) = f_T(t_1, t_2, \dots t_n | \theta) = \prod_{i=1}^n f_T(t_i | \theta).\label{eq:likelihood}
\end{equation}

Where $t_i$ are the sample observations for the node going between the two states of interest (for example, Unhealthy to Ready). We then optimize this function with respect to θ to get the best fitting parameters. Another complication arises when we have censored samples. What this means is that the FC does not give us the luxury sometimes to observe the samples. It loses patience with the organic recovery and takes some action after a certain time (say $x$). All we can say for these data points ($x_j$) is that the organic recovery took longer than $x_j$, nothing more and nothing less. If we want to incorporate these samples, we will need to include the probabilities of the organic recovery time exceeding these samples (not equal like we did with $t_i$). Hence, we must replace the PDF with the survival function ($S = 1-CDF$) making the likelihood function – 

\begin{equation}
\begin{split}
L(\theta; t_1, t_2, \dots t_n, x_1, x_2, \dots x_m)  \\
= \left(\prod_{i=1}^n f_T(t_i | \theta) \right)\left(\prod_{j=1}^m S_T(x_j | \theta) \right) \label{eq:likelihood}
\end{split}
\end{equation}

We can then optimize this for various distributions with respect to the parameters of those distributions (example, Lomax). 

Now, let’s turn to the question of regressing these models with features. Every time we observe any of the data points ($t_i$ or $x_j$ from the previous section), we also observe a plethora of information pertaining to that particular node. It has a certain hardware and software version, we know how many times it has failed in the past, etc. These are the features associated with the node. In the likelihood function described by equation (10), we ignored these features and considered only the organic and censored recovery times. How do we modify that function to take features into account? In the case of equation (10), we want to eventually get to the vector $\theta$ of parameters, which are the shape and scale. So, we need a way to go from the space of the feature vector ($f_i$) which has dimensionality, say, nx1 (including a bias term, so the number of features would in this case be (n-1)) to the shape and scale parameters (2x1) which make up the vector $\theta$. A matrix (say $W$) is used to transform this feature vector (with bias term) to the vector representing the shape and scale parameters of the distribution ($\theta$). So, if we pick the right dimensions for $W$, we can write – 

$$[\lambda, \kappa] = \theta_i = W f_i$$

\begin{figure}
  \includegraphics[width=\linewidth]{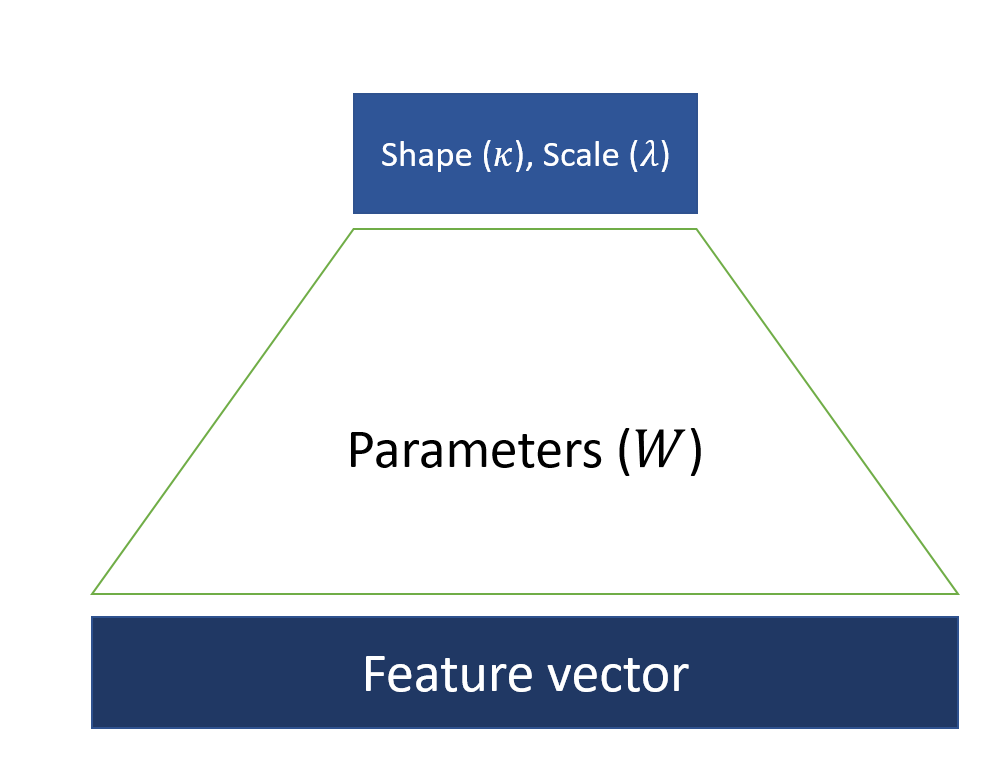}
  \caption{Feature vectors translated to shape and scale parameters.}
  \label{fig4}
\end{figure}

In other words, the matrix W transports vectors from the n-dimensional feature space (with constant bias) to the two-dimensional space of shape and scale parameters.

In practice, we want the shape and scale parameters to be positive, but the vector $W f_i$ is linear and so, not guaranteed to have positive components for all feature and parameter vectors. So, we apply sigmoid functions $(σ(x,a)=  a/(1+e^(-x) ))$ that constrain the resulting shape and scale between zero and an upper bound a. So, the modified equation becomes -

\begin{equation}
\left( \begin{array}{c}
		\kappa  \\
		\lambda 
	\end{array} \right) = \sigma \left(  \left( \begin{array}{ccc}
		w_{1,1} \dots w_{1,n}  \\
		w_{2,1} \dots w_{2,n}
		\end{array} \right)  \left( \begin{array}{ccc}
		f_{i1}  \\
		\vdots \\
		f_{in} 
		\end{array} \right) \right)
\end{equation}

Where,

\[W = \left( \begin{array}{ccc}
		w_{1,1} \dots w_{1,n}  \\
		w_{2,1} \dots w_{2,n}
		\end{array} \right) \]

And,

\[
f_i = \left( \begin{array}{ccc}
	f_{i1}  \\
	\vdots \\
	f_{in} \end{array} \right)
\]

The log likelihood from (10) will now be a function of the matrix $W$ –

\begin{equation}
	L(W) = \left(\prod_{i=1}^n f_T(t_i | f_i,W) \right)\left(\prod_{j=1}^m S_T(x_j | f_j,W) \right)
\end{equation}

Since log-likelihoods are a lot simpler to work with than likelihoods, let’s first take the logarithm of both sides to get the log-likelihood function.

\begin{equation}
	ll(W) = \sum_{i=1}^n ln (p(t_i | f_i,W))  + \sum_{j=1}^m ln (S_T(x_j | f_j,W) )
\end{equation}

Now, in order to optimize this function with respect to W, we need to find the gradient of the likelihood function above. Then, we can use simple gradient descent to get the optimal $W$ by simply picking a random starting W and moving along the gradient till we reach an optimum. In order to apply the chain rule and obtain the gradient of equation (13), we will first need to make it clearer. Let’s define a two-dimensional operator, $σ()$ that works on a two-dimensional vector like so: 

\[
\sigma(x) = \left( \begin{array}{ccc}
		\frac{U_1}{(1+e^{-x[1]})}  \\
		\frac{U_2}{(1+e^{-x[2]})}
		\end{array} \right)
\]

Where $U_1$ and $U_2$ are the upper bounds for the shape and scale parameters respectively. 

Then, equation (13) becomes,

\begin{equation}
	ll(W) = \sum_{i=1}^n ln (p(t_i | \sigma(W f_i) ))  + \sum_{j=1}^m ln (S_T(x_j | \sigma(W f_j)))
\end{equation}

Also, if $Wf_i=\alpha_i$, then

\[
\left( \begin{array}{c}
		\kappa  \\
		\lambda 
	\end{array} \right) = \theta = \sigma(\alpha)
\]

Taking derivative with respect to W for equation (14), we get the gradient –

\begin{equation}
\begin{split}
\nabla_W ll = \sum_i \left( \left(\frac{\partial ln (p)}{\partial \theta_i} \odot  \frac{\partial \theta_i}{\partial \alpha_i} \right) \left(\frac{\partial \alpha_i}{\partial W} \right)^T \right) \\
+ \sum_j \left( \left(\frac{\partial ln (S)}{\partial \theta_j} \odot  \frac{\partial \theta_j}{\partial \alpha_j} \right) \left(\frac{\partial \alpha_j}{\partial W} \right)^T \right)
\end{split}
\end{equation}

Since $Wf_i=\alpha$, we get $\frac{\partial \alpha}{\partial W}=f_i$ and since $\theta=\sigma(\alpha)$, we get $\frac{\partial \theta}{\partial \alpha}=\sigma' \alpha$, which is a vector of derivatives of the sigmoid operator. Substituting into equation (15), we get –

\begin{equation}
\begin{split}
\nabla_W ll = \sum_i \left( \left(\frac{\partial ln (p (t_i|\theta_i))}{\partial \theta_i}  \right) \circ \sigma^\prime(\alpha_i) \right)f_i^T \\
+ \sum_j \left( \left(\frac{\partial ln (S (x_j|\theta_j))}{\partial \theta_j}  \right) \circ \sigma^\prime(\alpha_j) \right)f_j^T \\
\end{split}
\end{equation}

And we can take the derivative of the sigmoid operator like so -

\[\sigma^\prime(\alpha) = \sigma(\alpha) \circ \left(  \left( \begin{array}{c}
		1  \\
		1
	\end{array} \right) -\sigma(\alpha) \right)\]

The $(\partial ln⁡(p(t_i |\theta_i))/(\partial \theta_i )$ and $(\partial ln⁡(S(x_j |\theta_j))/(\partial \theta_j )$ are the only parts that depend on the choice of our distribution and can be calculated easily for most distributions (Lomax, Weibull, Loglogistic, etc.). Using equation (16), we can easily calculate the gradient and perform gradient descent to get the optimal W that fits the data best. 

Though we can start with any random starting seed for $W$, the results become a lot more reliable when we start with –

\[
W = \left( \begin{array}{c}
		\alpha[1], 0 \dots 0  \\
		\alpha[2], 0 \dots 0
	\end{array} \right)
\]

And,

\[\alpha = \sigma(\theta)\]

Where $\theta$ is the vector of parameters that optimizes equation (13), the shape and scale parameters without features. This ensures that we start from the global θ which fits the whole data (without features) and move to customizing $\theta_i$ for each data point through $W$ from there.

For this to work, we must set the first element of every $f_i$ to 1 to account for the constant term.

\subsection{Effect of using features on savings}
In figure 7 (below), we show how the savings and log likelihood move in tandem across the iterations of our gradient descent (starting with the global optimal threshold) on some data. This demonstrates that we can expect better and better savings as we fit the data better. The likelihood and savings start at the level of the model without features (due to the way we choose initial parameters). We get a ~17\% increase in the log likelihood when using cluster level features (with a further ~1\% increase when we add node level features as well).

\begin{figure}
  \includegraphics[width=\linewidth]{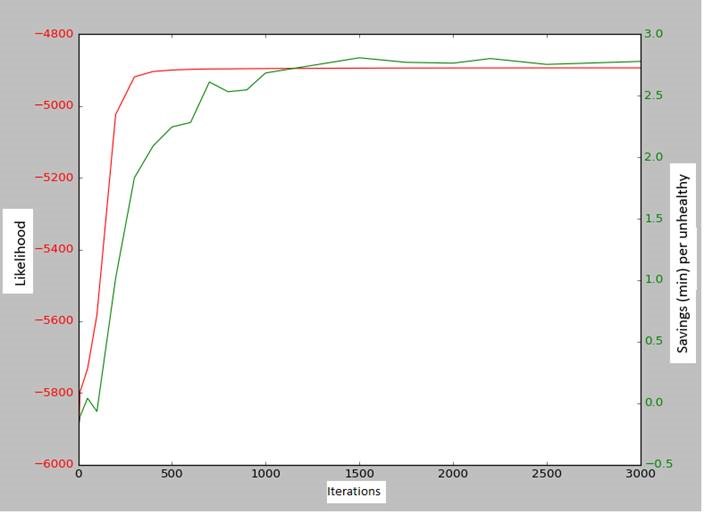}
  \caption{Demonstration that the savings from a model improves with log likelihood. The x-axis represents the number of iterations that the gradient descent has run through. The red curve is the log likelihood and the green curve is the savings per instance of Unhealthy node (both of which improve as gradient descent progresses). The starting point here was the model without features, so using features can definitely produce greater savings.}
  \label{fig8}
\end{figure}

\section{Measuring the performance of the optimal thresholds in production }

Now that we have our optimal thresholds deployed, we need to objectively measure how much they're moving the needle, which is the KPI (key performance indicator) everyone in Azure cares about: virtual machine downtime. And it is very important that this be done in a way that doesn't depend on the assumptions of the model. In most cases, we end up lowering the thresholds to different degrees. You can see from figure 6 below the challenge this poses in terms of measuring the model. When we lower it from the green level to the yellow level,

\begin{itemize}
\item{For the orange dots, we will end up unnecessarily rebooting them early when they would have recovered had we waited a little longer. So, will probably incur more downtime.}
\item{For the green dots, they were rebooting at the current threshold anyway. So, it is advantageous to reboot them as soon as possible (if they’re going to reboot anyway, why wait?). For these, we end up saving downtime (by the exact amount that the threshold was lowered).}
\item{However, once the threshold is lowered, we can’t distinguish between the orange and green dots. Both get censored at the new, lower threshold.}
\end{itemize}

\begin{figure}
  \includegraphics[width=\linewidth]{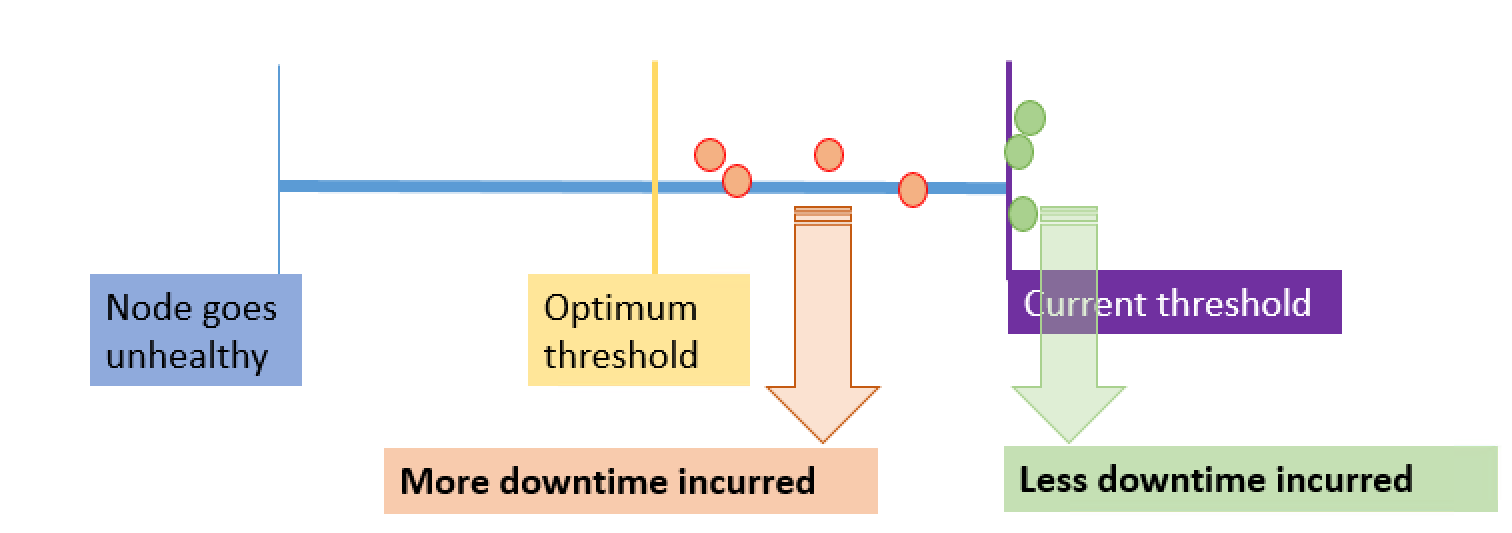}
  \caption{Consequences of lowering thresholds on data censoring. The x-axis is the time we wait for a node to recover when unhealthy. The green dots represent nodes that are being power cycled at the current threshold. So, we will certainly benefit from rebooting them earlier. The orange dots are nodes that went back to Ready between the (lower) optimal threshold and the current threshold. For these, we might have incurred more downtime since we would have rebooted them unnecessarily. Once the threshold is lowered, it will be impossible to distinguish the orange points from the green ones since they will all be rebooted. This is the reason we need randomization. We use the optimal threshold not always, but a certain percentage of the time.}
  \label{fig:fig5}
\end{figure}

We could use the distributional assumption we used for modelling the time to organic recovery to estimate how many of the censored dots fell into the green and orange categories, but then we would be using the assumptions of the model to measure the performance of the model. 
To address this, we put together mechanisms for un-biased experimentation in place. Instead of blindly using the recommended threshold all the time, the FC does the equivalent of tossing a coin with a certain probability of “heads” each time it is to pick a threshold. It then uses the optimal thresholds if it gets “heads”, but sticks to the old threshold (of 10 minutes) if it gets “tails”. This probability with which the FC selects the optimal threshold is configurable (it is currently set at 35\% across all of production). This setup is inspired by A/B testing that is prevalent in the web world (a new feature is rolled out only not at once but gradually to accommodate measurement in performance). 
Now, we will get from the model in production, instances of unhealthy node recovery when we used the optimal threshold (treatment group) and when we used the old threshold (control group). We can then join these logs with the stream for virtual machine downtime and compare the average downtime. By calculating the means and standard deviations of the downtimes across the two groups, we can get estimates of statistical significance (p-values) using the two-sample t-test. More on this in the Results section.

\section{Deployment mechanism for the optimal thresholds}
As you might imagine, changing all these settings in a complicated system like Azure is a scary business with plenty of scope for nightmarish consequences if things go wrong. If our Unhealthy node recovery threshold is too low, for example, we might see a huge number of nodes being unnecessarily rebooted, causing all the VMs to go down. So, we should have safe deployment mechanisms in place where any thresholds are thoroughly inspected and vetted and get deployed gradually so that problems can be potentially caught before too much harm is done. We tried to address these concerns in the deployment pipeline we put together. For now, the FC is consuming only the Unhealthy recovery thresholds, which are calculated per cluster (collection of a few hundred to a thousand nodes, generally uniform in hardware). 

The big data platform used within Microsoft is called Cosmos which is capable of running map-reduce jobs. We can write custom reducers in the C\# programming language that enables us to run optimization algorithms on large data sets (like the node transition data described above). As you can imagine, map-reduce jobs are bulky and slow. For fast querying, we have another data store called Kusto that persists all data mostly in memory and is hence suitable for quick querying and slicing of the processed results.

Using these components, we have set up the following data processing pipeline for getting the optimal thresholds to production periodically:

\begin{itemize}
\item{The FC logs all possible data pertaining to the state transitions and other things that might affect them along with the values of the thresholds and settings it is using. These logs are pushed to Cosmos where data science teams like ours can access this data.}
\item{Our Cosmos jobs then read this data daily and run our models on it. This allows for periodic calculation of the thresholds (once a day), allowing them to adapt to the changing data.}
\item{The optimal thresholds are then copied over to Kusto from Cosmos. This allows for quick querying for debugging purposes, and also allows a PowerShell script to read their latest values and copy them over to CSV files, which are called “model files”. The PowerShell script then generates a pull request for these files so they can be reviewed by members of the insights and Fabric teams.}
\item{These files then get checked in to source control and flow as part of the deployment process to production. The deployment train generally takes 2-3 weeks to get to production and then, we can enable them and start monitoring the performance as described in the previous section.}
\item{Once the FC starts acting based on our thresholds and logging the resulting behavior, the logs flow once again to Cosmos and the cycle repeats.}
\end{itemize}

These steps are illustrated in the flowchart below (figure 7). The process forms a feedback loop where we can monitor the performance of our models over every cycle and adjust as required to continue improving.

\begin{figure}
  \includegraphics[width=\linewidth]{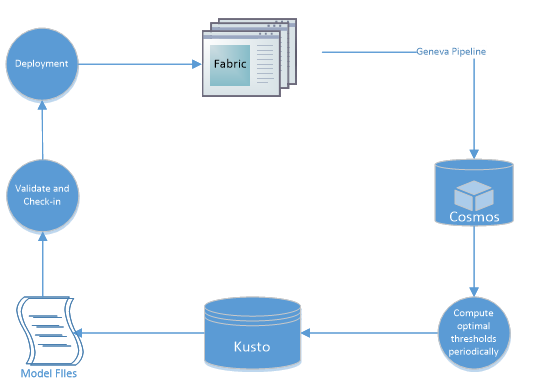}
  \caption{Deployment pipeline for pushing thresholds to the FC}
  \label{fig6}
\end{figure}

\section{Moving from one to multiple thresholds within the state machine}

The process of finding the optimal threshold outlined in the section 5, only holds for the simplest of transition matrices. In particular, we assumed the intervention cost $C_{int}$ was not affected by the waiting threshold, $\tau$. So, we were first able to find $C_{int}$ and then use its estimated value to find the optimal waiting threshold,  $\hat{\tau}$. In other words, we can calculate them sequentially. This is certainly a good assumption if, once we send the node to PoweringOn, there is no chance for the node to come back to Unhealthy (which happens to be true for our data). And not just directly, but via any longer path (ex: PoweringOn to Raw to Unhealthy).  For example, this would be a good assumption for the following matrix of transition probabilities (since PoweringOn always goes straight to Ready with no chance of a path ever leading back to Unhealthy) –

\begin{equation}
\bordermatrix{ & Unhl & PowOn & Rdy \cr Unhl & 0 & P(T > \tau) & P(T < \tau) \cr PowOn & 0 & 0 & 1 } \qquad
\end{equation}

If it does, however, then $\tau$ will end up affecting the time it takes to go from PoweringOn to (eventually) Ready as well. So, we can no longer compute $C_{int}$ and  $\hat{\tau}$ sequentially.  We can construct such a matrix by inserting a finite probability ($p$) for the node going from PoweringOn back to Unhealthy. The matrix will now look like -

\begin{equation}
\bordermatrix{ & Unhl & PowOn & Rdy \cr Unhl & 0 & P(T > \tau) & P(T < \tau) \cr PowOn & p & 0 & (1-p) } \qquad
\end{equation}

Also, let’s assume that the time it takes to go from PoweringOn back to Unhealthy is always $B$ (time to go back). The matrix of average times to make transitions will look like –

\begin{equation}
\bordermatrix{ & Unhl & PowOn & Rdy \cr Unhl & 0 &  \tau & E[T|T<\tau] \cr PowOn & B & 0 & C_{int} } \qquad
\end{equation}

In figure 8 (above), we show how the intervention cost ($C_{int}$) changes with the Unhealthy threshold ($\tau$) for different values of $p$. For the real data, $p$ is very close to zero and so, the dependence is flat and we can ignore it.

\begin{figure}
  \includegraphics[width=\linewidth]{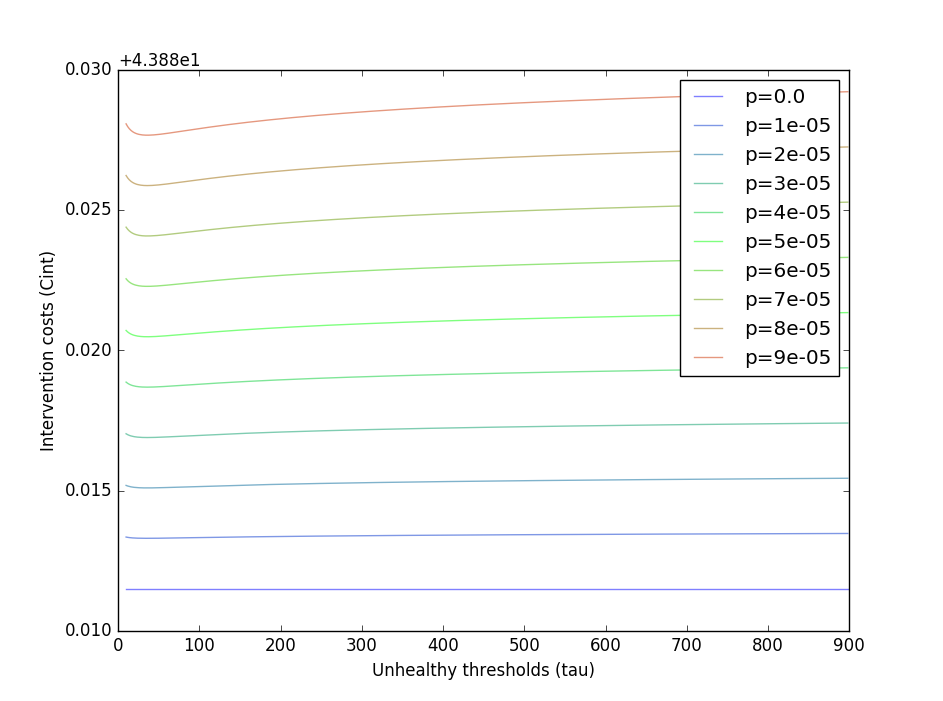}
  \caption{Demonstration of how the Intervention Cost ($C_{int}$) changes with the Unhealthy threshold ($\tau$) for different values of p as defined in the text.}
  \label{fig7}
\end{figure}

If $p$ is not too small, the  $\hat{\tau}$ that satisfied equation (5d) will probably not be optimal, since we ignore that it doesn’t impact the reboot time. If we set it too small now for example, there is a greater risk of nodes cycling between Unhealthy and PoweringOn, hence incurring larger downtime. So, we should expect the new optimal $\tau$ to be somewhat larger than the  $\hat{\tau}$ that satisfies (5d). To obtain it, we'll need to solve equation (8) for t, pick the first element from this vector (which corresponds to the average time to go from Unhealthy to Ready counting transitions through intermediate states), and optimize it with respect to $\tau$. It might still be possible to solve this optimization problem in closed form, but as these transition matrices become larger and more intricate, it won’t be. In those cases, we might need to use numerical techniques.

So far, we considered the threshold dictating the movement from Unhealthy to Ready. It is not hard to imagine that there might be other thresholds dictating similar movements between other pairs of states. For example, there is a very similar threshold that takes the node from PoweringOn to Ready. 

When the node is rebooting, the FC can’t wait forever for it to come back up (just like with Unhealthy). And if it doesn't, the FC sets the node to HumanInvestigate, moves the VMs off it to a healthy node and then various diagnostics are performed on the node. And when we have many of these thresholds, it is easy to imagine that they might affect each other. 
So, due to the relationships between them, optimizing them together is superior to optimizing each one in isolation in order to get the largest savings possible. Also, coming up with a framework for optimizing any number of thresholds affecting the transitions within a state machine leads us towards a generic method that can be ported to any problem that can be expressed in terms of states (which probably has coverage over problems from a wide range of domains). We will explore a generic method to solve these kinds of problems, though it's not something we've tested yet in a production setting.

Let's consider a simple scenario. Let's say there are two thresholds, one is for deciding when to PowerCycle a node when it goes Unhealthy ($\tau_1$) (this was $\tau$ earlier) and the other decides when to HI a node when it is rebooting (in the state PowerCycle) ($\tau_2$). The organic recovery duration of going from Unhealthy to Ready is a random variable denoted by $T_1$ and that of going from PoweringOn to Ready is $T_2$ (both random variables to be modelled by some distributions). 

Like before, let’s say that when the node is in PoweringOn, it might go back to Unhealthy (say, due to a bug in the FC). The probability of going from PoweringOn to Unhealthy is $p$, and it always takes a certain amount of time (call it $B$). If $p=0$, this leads to the standard case. So, the matrices of transition probabilities and average times look like - 

Transition probabilities:

\small
\begin{equation}
\bordermatrix{ & Unhl & PowOn & HI & Rdy \cr  & 0 &  P(T_1 > \tau_1) & 0 & P(T_1 < \tau_1) \cr 
						        & p & 0 & (1-p)P(T_2 > \tau_2)  & (1-p)P(T_2<\tau_2) \cr
						            & 0 & 0 & 0 & 1 \cr
						             & 0 & 0 & 0 & 0} \qquad
\end{equation}

\begin{equation}
\bordermatrix{ & Unhl & PowOn & HI & Rdy \cr  & 0 &   \tau_1 & 0 & E[T_1 | T_1 < \tau_1] \cr 
						        & B & 0 & \tau_2  & E[T_2 | T_2 < \tau_2] \cr
						            & 0 & 0 & 0 & C_{HI} \cr
						             & 0 & 0 & 0 & 0} \qquad
\end{equation}

As discussed, the average times for going from each of the states to Ready can be represented as a vector t which satisfies –

\begin{equation}
(\mathbbm{I} - \mathbf{Q})\mathbf{t} = (\mathbf{P} \odot \mathbf{\mathcal{T}}) \cdot \mathbf{1}, \label{eq:cost}
\end{equation}

Where $Q$ is the same as $P$, but with the row and column corresponding to the Ready state removed. We then read off the element of $t$ corresponding to Unhealthy and try to optimize it with respect to $\tau_1$ and $\tau_2$. There is no convenient closed form for this optimal vector $[(\hat{\tau_1} ),(\hat{\tau_2} )]$. However, we can numerically calculate the gradient and perform gradient descent. 

When $p=0$, we expect the result to be the same as optimizing $\tau_2$ first and then using the optimal expected downtime for the reboot cost and use it to optimize $\tau_1$. And indeed, when we use simple Lomax distributions for $T_1$ and $T_2$, we do see this. The Newton Raphson routine gives us the expected vector of thresholds. When $p>0$ however, we do expect the PoweringOn threshold ($\tau_2$) to be slightly lower and the Unhealthy threshold ($\tau_1$) to be slightly higher (because the model will try to avoid Unhealthy to PoweringOn back and forth loops, so giving organic recovery while in Unhealthy more of a chance and going to HumanInvestigate are more attractive options now). And this is indeed what we get when we run the optimization. 

\section{Results}
We deployed models that optimize the Unhealthy to PoweringOn threshold using cluster level features fitted to the LogLogistic distribution and PoweringOn to HumanInvestigate as well as Booting to PoweringOn thresholds at the cluster level (one model per cluster) using the same distribution. We have managed to reduce the customer downtime by 3\% for the Unhealthy scenario and 4-5\% for the PoweringOn and Booting scenarios. This represents significant savings and a better experience for our customers.

\section{Conclusion}
In this paper, we explored methods to find the optimal time to wait for something in the context of Azure thresholds and settings. This can easily be extended to any software component that relies on APIs exposed by other software components. There is always a dilemma of how long one should wait. For a non-software scenario, consider someone waiting for a bus everyday when they know they can walk to their destination as well. How long should they wait for the bus before giving up and starting to walk? Since the tools needed for scenarios like these like censored maximum likelihood estimation are pretty generic, we also published an open-source library that implements these generic methods and demonstrates how to apply them to such optimum waiting problems: \url{https://github.com/ryu577/survival}.

\begin{appendices}
\section{Derivation of expected downtime greater than $t$ for Lomax}
In §3.2, we promised a derivation for the expected value of the Lomax random variable given that it is greater than (or less than) $t$. This quantity was used in the second part of the expression for expected downtime given in equation (1). We evaluate this expression here - 

We know, 
$$\mathbbm{E}[X|X>t] = \int_t^\infty x.\frac{f_X(x)}{\mathcal{S}_X(t)} dx$$
Note that to get the truncated distribution, we needed to divide the PDF by the survival function. This is the only way to ensure that the truncated density integrates to one as all distributions must. Since the survival function doesn't depend on $x$, we can factor it out of the integral.

$$\mathbbm{E}[X|X>t].\mathcal{S}_X(t) = \int_t^\infty x.f_X(x) dx$$

Substituting the density for the Lomax distribution we get for the R.H.S.
$$ \int_t^\infty \frac{\lambda.\kappa.x}{(1+\lambda.x)^{\kappa+1}} dx $$

In order to solve this, we recall the formula for integration by parts - 
$$ \int u dv= uv - \int vdu$$

We note that - 
$$d\left( \frac{1}{(1+\lambda.x)^\kappa}\right) = -\frac{\lambda.\kappa}{(1+\lambda.x)^{\kappa+1}}dx$$

So, let 
$$v = \frac{1}{(1+\lambda.x)^{\kappa}}$$
and 
$$u = x$$

\begin{multline}\Rightarrow \int_t^\infty f_X(x)dx = \int_t^\infty -d\left( \frac{1}{(1+\lambda.x)^\kappa}\right)x \\ = -\left[\frac{x}{(1+\lambda.x)^\kappa}\right]_t^\infty + \int_t^\infty \frac{dx}{(1+\lambda.x)^\kappa}\end{multline}

$$ = \frac{t}{(1+\lambda.t)^\kappa} + \frac{1+\lambda.t}{\lambda.(\kappa-1).(1+\lambda.t)^\kappa}$$

Dividing by the survival function,
$$\mathcal{S}_X(t) = \frac{1}{(1+\lambda.t)^\kappa}$$
We get - 
$$\mathbbm{E}[X|X>t] = \frac{\kappa}{\kappa-1}.t + \frac{1}{\lambda.(\kappa - 1)}$$

We have hence derived equation (8). Note that $\kappa$ must be more than one for this expression to hold. If $\kappa < 1$, the integral above blows up and tends to $\infty$. Basically, the distribution becomes so heavy tailed that it lacks a first moment (I know there is a ``Yo mama" joke there but I can't put my finger on it).

\section{Derivation of the closed form solution of the expected downtime as a function of $\tau$}
We derived in the Appendix A the form for $\mathbbm{E}[X|X>t]$. We can now use the fact that - 

$$\mathbbm{E}[X] = P(X<\tau).\mathbbm{E}[X|X<\tau] + P(X>\tau).\mathbbm{E}[X|X>\tau]$$

to find $P(X<\tau).\mathbbm{E}[X|X<\tau]$ which was the first part of the expected downtime $\mathbbm{E}[T]$ in equation (1). 

We get - 
$$\mathbbm{E}[X|X<\tau].P(X<\tau) = \mathbbm{E}[X] - \mathbbm{E}[X|X>\tau].P(X>\tau)$$

$$ = \frac{1}{\lambda (\kappa-1)} -  \left[ \frac{1}{\lambda (\kappa-1)} + \tau \frac{\kappa}{\kappa - 1} \right]    (1+\lambda.\tau)^{-\kappa}$$

Note that this expression will give a valid result even if $\kappa < 1$ unlike the expression in Appendix A. This is because the integral is bounded. We can now substitute this into equation (1) and we will get the expression for $\mathbbm{E}[T]$ corresponding to the Lomax distribution given in  equation (10).

\end{appendices}

\end{document}